\documentclass{article}
\makeatletter
\renewcommand{\@latex@error}[2]{}
\makeatother

\usepackage{arxiv}

\usepackage[utf8]{inputenc} 
\usepackage[T1]{fontenc}    
\usepackage{hyperref}       
\usepackage{url}            
\usepackage{booktabs}       
\usepackage{amsfonts}       
\usepackage{nicefrac}       
\usepackage{microtype}      
\usepackage{lipsum}		
\usepackage{graphicx}
\usepackage{natbib}
\usepackage{xspace}
\usepackage{doi}
\usepackage{graphicx}
\usepackage{rotating} 
\usepackage{booktabs}
\usepackage{latexsym}
\usepackage{arabtex}
\usepackage{utf8}
\usepackage{amsmath}
\usepackage{comment}
\usepackage{abstract}
\setcode{utf8}
\usepackage{hyperref} 
\newcommand{\undertitle}[1]{\textit{#1}}

\title{FASSILA: A Corpus for Algerian Dialect Fake News Detection and Sentiment Analysis}

\author{ \href{https://orcid.org/0009-0001-8903-3137}{\includegraphics[scale=0.06]{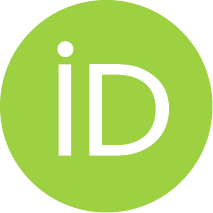}\hspace{1mm}Amin ABDEDAIEM} \\
	Department of Mathematics and Computer Science\\ Ahmed Draia University\\
	Adrar,Algeria \\
	\texttt{aminabdedaiem@gmail.com} \\
	\And
	\href{https://orcid.org/0000-0001-8793-2465}{\includegraphics[scale=0.06]{orcid.pdf}\hspace{1mm}Abdelhalim Hafedh Dahou} \\
	GESIS – Leibniz-Institute for the Social Sciences\\
	Cologne, 50667, Germany \\
	\texttt{abdelhalim.dahou@gesis.org} \\
\And
\href{https://orcid.org/0000-0003-2760-5547}{\includegraphics[scale=0.06]{orcid.pdf}\hspace{1mm}Mohamed Amine Cheragui}\\
	Department of Mathematics and Computer Science\\ Ahmed Draia University\\
	Adrar,Algeria \\
	\texttt{m\_cheragui@univ-adrar.edu.dz} \\
	\And
 	\href{https://orcid.org/0000-0003-1793-9615}{\includegraphics[scale=0.06]{orcid.pdf}\hspace{1mm}Brigitte Mathiak} \\
	GESIS – Leibniz-Institute for the Social Sciences\\
	Cologne, 50667, Germany \\
	\texttt{brigitte.mathiak@gesis.org} \\
}

\renewcommand{\headeright}{Technical Report}
\renewcommand{\shorttitle}{\textit FASSILA}

\hypersetup{
pdftitle={A template for the arxiv style},
pdfsubject={q-bio.NC, q-bio.QM},
pdfauthor={David S.~Hippocampus, Elias D.~Striatum},
pdfkeywords={First keyword, Second keyword, More},
}

\begin{document}
\maketitle
\begin{abstract}
In the context of low-resource languages, the Algerian dialect (AD) faces challenges due to the absence of annotated corpora, hindering its effective processing, notably in Machine Learning (ML) applications reliant on corpora for training and assessment. This study outlines the development process of a specialized corpus for Fake News (FN) detection and sentiment analysis (SA) in AD called FASSILA. This corpus comprises 10,087 sentences, encompassing over 19,497 unique words in AD, and addresses the significant lack of linguistic resources in the language and covers seven distinct domains. We propose an annotation scheme for FN detection and SA, detailing the data collection, cleaning, and labelling process. Remarkable Inter-Annotator Agreement indicates that the annotation scheme produces consistent annotations of high quality. Subsequent classification experiments using BERT-based models and ML models are presented, demonstrate promising results and highlight avenues for further research. The dataset is made freely available on GitHub \footnote{\url{https://github.com/amincoding/FASSILA}} to facilitate future advancements in the field.
\end{abstract}


\section{Introduction}

Building a corpus become an important topic in natural language processing (NLP) and especially for low resource languages (ex: AD), due to the importance that the corpus plays in the development of several tools, such as: Machine Translation ~~\cite{babaali2022survey}, Part of speech tagging ~~\cite{chiche2022part}, Named entities recognition ~~\cite{jarrar2022wojood}, etc . in particular with the emergence of techniques based on statistics, machine learning and deep learning.  Who exploits this mass of information to develop, train and evaluate models.
However, building a corpus is not an easy task ~~\cite{Bakari2016AQAWebCorpWF}; it is extremely time-consuming and requires a lot of work, for the good reason that the volume and quality of the corpus are two important parameters. Despite the recent emergence of techniques that consume fewer resources, such as few-shot learning ~~\cite{setfit}. 

Over the last few years, a lot of studies in NLP have focused on languages or variants of languages called low resources ~~\cite{mengoni2023special}. This change of direction is mainly due to the emergence of social media such as Facebook, Twitter, RenRen, LinkedIn, Google+, and Tuenti, as a means of communication where people exchange messages and comments. In Arab countries, these exchanges are carried out using variants of the Arabic language known as dialects. 
The boom in the use of social media has not only provided raw data for building a corpus, but has also led to the emergence of other topics, such as detecting Fake News and Sentiment Analysis.     
also known as opinion mining,Sentiment Analysis is the field of study that analyses people's opinions, feelings, evaluations, attitudes and emotions towards entities and their attributes expressed in written text ~~\cite{zhao2016sentiment}.
On the other hand, the detection of Fake News is aimed at combating a phenomenon that has grown with the growth of social media, given the amount of information circulating and which can have an impact on a number of areas, whether social, economic or even political ~~\cite{nevado2023analysis}.

The aim of our study is to contribute to the development of resources for the AD, through the building of a corpus dedicated to the detection of Fake News and the Sentiment Analysis. To this, we add a series of experiments on Fake News based on machine learning techniques such as: Support Vector Machines (\textit{SVM}), Logistic Regression (LR), Decision Trees (\textit{DT}); and transformer-based models including:  AraBERTv02, \textit{MarBERTv2}, and \textit{DziriBERT}.

Our study endeavors to enrich the resources available for the Algerian Dialect by constructing a specialized corpus tailored for the detection of Fake News and Sentiment Analysis. This initiative stems from the recognition of the scarcity of linguistic resources in Algerian Dialect and the critical need to empower computational linguistics with dedicated datasets. Through the creation of this corpus, we aim to facilitate more robust and nuanced analyses of Fake News and Sentiment Analysis within the Algerian Dialect discourse, thereby advancing research and applications in computational linguistics for this understudied language.
\label{sec:introduction}

In the landscape of linguistics, the Algerian Dialect stands as one of the most low-resource languages, lacking an official writing style or a sustained repository of data. This scarcity poses significant challenges across linguistic applications, ranging from translation models to the detection of FN. Without a standardized corpus, Algerian Dialect remains largely untapped in computational linguistics.

The imperative to create a dedicated corpus for Fake News detection and Sentiment Analysis in Algerian Dialect arises from this dearth of linguistic resources. Establishing such a corpus not only addresses the pressing need for linguistic data but also unlocks avenues for advancing machine learning methodologies tailored to AD's unique linguistic nuances.

This study aims to achieve the followings:
\begin{itemize}

\item Development of the worlds first specialized corpus tailored for Fake News detection and Sentiment Analysis in the Algerian Dialect in parallel.
\item Establishment of a corpus consisting of 10,087 sentences, encompassing over 15,561 unique words in AD, addressing the significant dearth of linguistic resources in the language.
\item Testing out corpus to prove its fine tuning capabilities and strong maintainability.
\item testing GPT-4's paraphrasing and translation abilities and using them once proven to be reliable.
\end{itemize}

The rest of the paper is laid out as follows: Section 2 gives an overview of the AD. Section 3 presents the related works regarding the detection of Fake News and Sentiment Analysis in the AD; Section 4 presents how we build our corpus FASSILA; Section 5 describes the experimental Setting. Experimental results are discussed in Section 6, and we conclude and present future work in Section 7.

\section{Related Work}\label{sec:relatedwork}
In the realm of Fake News detection and Sentiment Analysis within the Algerian Dialect context, prior studies have explored various methodologies and techniques. ~~\cite{guellil2018sentialg} proposed an approach for Sentiment Analysis in AD, achieving F1-scores of up to 72\% and 78\% for Arabic and Arabizi internal test sets, respectively. However, limitations exist in the broad applicability of their method across diverse linguistic contexts, were as ~~\cite{rahab2021sana}introduced a classification method for Arabic comments from Algerian Newspapers, attaining notable F-scores with SVM and NB classifiers. Nevertheless, the study's scope may constrain its generalizability beyond newspaper comments.~~\cite{abdelli2019sentiment} employed supervised methods like LSTM and SVM for SA, yet their focus on supervised techniques might overlook potential benefits of unsupervised approaches. Chader et al. ~~\cite{chader2019sentiment} introduced a supervised Sentiment Analysis method on Arabizi AD, showcasing promising results with SVM. However, the study's emphasis on preprocessing techniques may not fully address the linguistic diversity present in AD. ~~\cite{mazari2022sentiment} conducted an experimental study on ML and DL algorithms for SA, demonstrating accuracies ranging from 64.11\% to 84.21\%. Yet, their comparison of multiple models may overlook nuanced performance differences across linguistic contexts. 
~~\cite{Rumor_Stance_Classification} employed three DL models, mBert, XLM-Roberta, and AraBERT, to scrutinize a political rumor regarding the health of the Algerian President. Their dataset comprised 3,147 comments, categorized into six classes: "Support", "Deny", "Query", "Comment", "Positive Comment", and "Negative Comment". AraBERT outperformed other models, achieving an F1-score of 53\%, surpassing mBert 45\% and XLM-Roberta 38\%. ~~\cite{arabizi} introduced a rumor detection method for Algerian Arabizi, leveraging various classification models (LSTM, GRU, and CNN) and textual representations (Word2vec, bag of ngrams, and ELMo). Their dataset comprised 9,528 comments, with LTSM model demonstrating more than a 10\% improvement in performance using Word2vec and ngrams bag representations. Lastly, ~~\cite{abdedaiem2023fake} In a prior work of ours, we conducted a comparative study between SetFIT and fine-tuning methods, revealing SetFIT's superiority in Fake News detection in low-resource language contexts like AD. However, the limited dataset used may warrant further investigation into broader linguistic contexts. In addressing these limitations, our paper aims to contribute a comprehensive approach to Fake News detection and Sentiment Analysis in AD, leveraging insights from previous studies while expanding the scope to accommodate the linguistic diversity inherent in Algerian Dialect discourse.
\section{Corpus Construction}
In this section, we provide a detailed description of the methodology used to compile the FASSILA dataset and present its composition. The
collection of the data was based on the most used social media platforms used in Algeria, including Facebook and YouTube, in addition to pre-existing datasets written in MSA including Khouja corpus ~\cite{Khouja}, ANT corpus ~\cite{Chouigui_2017}, and lastly a set of hand-made and paraphrased sentences that meets the requirement of both tasks (which contains news and sentiments). 

\begin{figure}[h]
\centering
\includegraphics[scale=0.4]{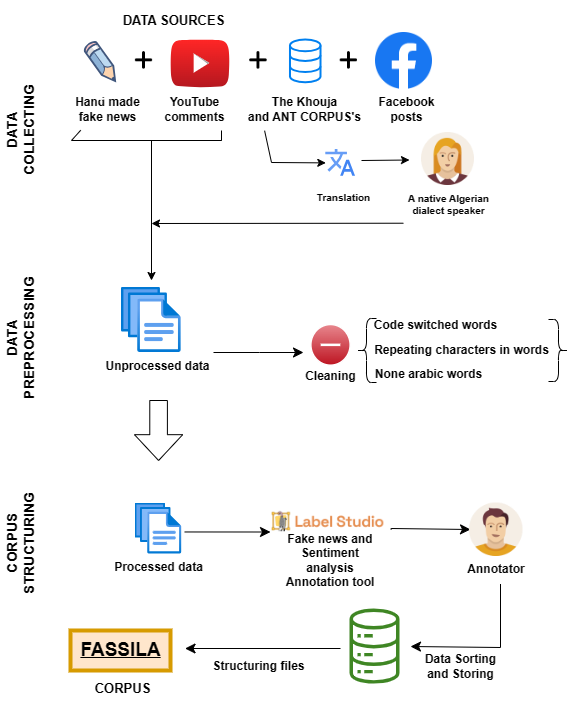}
\caption{Building the FASSILA Corpus.}
\label{fig:31}
\                                     
\end{figure}

The first step was gathering the data using tools like the YouTube API 3rd version as well as simple web scraping algorithms using Beautiful-soup library \footnote{https://beautiful-soup-4.readthedocs.io/en/latest/}, then we cleaned the data from any none Arabic letters, email, and code-switched  words (foreign words written in Latin characters).
After that, we used label-studio to annotate the unlabeled data, and after that the corpus becomes ready in a well presented schema (source, category, text, label) as shown in Figure \ref{fig:31}.

\subsection{Data Collection}

The FASSILA Corpus is built using data from various sources including YouTube, Facebook, handmade sentences, paraphrased sentences, and manually-translated sentences from the khouja dataset ~\cite{Khouja} and ANT-v1  corpus ~\cite{Chouigui_2017}. The process of gathering data for the research involved the utilization of the YouTube APIv3, a powerful tool provided by Google\footnote{\url{https://developers.google.com/youtube/v3)}} . This API allowed for the extraction of relevant YouTube comments that contained information with news or sentiment. By leveraging this API, a significant amount of data was collected, providing a valuable resource for analysis and investigation. For corpus diversity, we focused on the information areas that contain more news and users feedback and opinion, including worldwide politics, health, and five pertaining to the Algerian community - car prices, tourism, accidents, eCommerce, and the national football team's 2022 World Cup qualification in Qatar all if which has been collected in June 2023. For the fake news task, we did not set a specific time frame for data selection, the categories tackled in this corpus were current buzz topics in Algeria at the time of the corpus's creation. This approach aimed to maximize the existing data due to the Algerian Dialect being a low-resource language.
\subsection{Data Cleaning}
After collecting the comments, we carried out several
pre-processing steps. starting with removing special characters, such as punctuation marks, emails, random numbers, and repetitions. After removing special characters, the next step involved normalizing the text. This process aimed to standardize the textual data for consistency and uniformity throughout the corpus. One method involved converting all text to Arabic characters, which helped mitigate potential discrepancies arising from variations in the writing style. In our case, we transformed all the comments writing style to the Algerian dialect style using just Arabic characters as illustrated in the table \ref{tab:writing_styles}.
\begin{table*}[h]
\centering
\begin{tabular}{cc}
\hline
\textbf{Writing Cases} & \textbf{Algerian Sentence}  \\ \hline
Algerian dialect         & \<هاذاك الحادث صرا بسباب كاميو نتاع ايسونس ماشي سيما>   \\
Arabizi Style &   hadak lhadith sra bsbab camio ntaa essonss mashi sima \\ 
MSA          & \<ذاك الحادث وقع بسبب شاحنة نقل بترول وليس أسمنت>  \\ 
English translation          & That accident happened because of an oil transportation truck and not cement  \\ \hline
\end{tabular}
\caption{Examples of writing the Algerian dialect in different styles on social media, illustrated with a sentence from the FASSILA Corpus.}
\label{tab:writing_styles}
\end{table*} 

However, we acknowledged that Algerian Dialect often contain words in MSA due to their derivation from MSA, which preserves their originality. Such sentences were also considered for inclusion in the corpus. The included YouTube comments were only the sentences containing any news or sentiment information related to the information areas mentioned above. This decision was made to capture diverse sources of information and opinions in the Algerian dialect. In addition to that, we excluded irrelevant content types, such as advertisements, product reviews, or unrelated personal stories, or any kind of parody or extremism sentences. This helped ensure that our corpus only contained data relevant to the tasks.

\begin{figure}[h]
\centering
\includegraphics[scale=0.4]{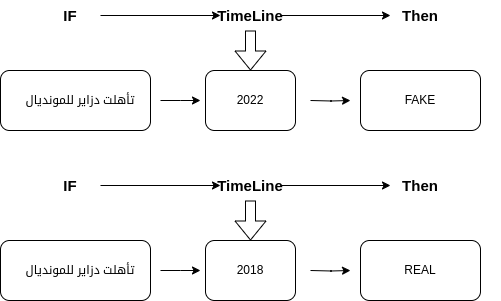}
\caption{Example of temporal consistency.}
\label{fig:timeline}
\                                     
\end{figure}




\subsection{Data augmentation}

In our research, we leveraged both paraphrasing and translation as data augmentation techniques. For paraphrasing, we utilized GPT-4 to generate rephrased sentences from the original ones, aiming to diversify the dataset while retaining the original meaning. However, we encountered a challenge where the paraphrased sentences tended to drift towards the Moroccan dialect instead of the desired Algerian dialect, which was also proved in ~\cite{babaali2024breaking}. 

To mitigate this issue, we implemented a control mechanism by providing a specific prompt to GPT-4 ~\cite{openai2023gpt4} as presented in \ref{app:prompts}. The prompt instructed the model to paraphrase the sentences while ensuring that the dialect remained consistent with the Algerian dialect and did not shift towards the Moroccan dialect. If the model mistakenly generated a sentence in the Moroccan dialect despite the prompt, it was returned unchanged with a '\#' appended at the end.

In addition to paraphrasing, we also employed translation from MSA to the Algerian dialect for the khouja corpus claim side that comprised of 4,547, as another data augmentation technique using GPT-4 after an automatic and manual assessments. The detailed process of translation mechanism is provided in Appendix \ref{app:breakdownT}. Similar to paraphrasing, we applied the same process where we translated the sentences trailing with '\#' manually which made the corpus contains only northern (algerois) dialect and southern adrarian dialect. Around 65\% of the dialect used in the translation was the Adrarian (A city in south of Algeria) and the remainder is mainstream Algerian dialect which are despite the fact that they might sound different but the majority of the differences are pronunciation and vocabulary, and wouldn't harm the corpus in any ways because they are still Algerian dialect.

We have utilised translation using GPT-4 
by incorporating these control mechanisms, we ensured that both paraphrased and translated sentences retained the linguistic characteristics of the Algerian dialect, thereby enriching the training dataset with diverse yet dialect-consistent data instances. This approach enhanced the robustness and effectiveness of our natural language processing models, enabling them to better understand and generate text in the Algerian dialect.

\subsection{Annotation and Labeling}
In this section, we discuss the annotation and labeling process for building the FASSILA corpus, which is designed for FN detection and SA in the AD. The corpus comprises four sources of sentences: 

\begin{itemize}
    \item Sentences translated from the Khouja. 
    \item Manually created sentences based on current events. 
    \item Sentences extracted from YouTube comments and Facebook pages.
    \item Paraphrased sentences.
\end{itemize}

Each type of sentence underwent a distinct annotation and labeling process to ensure the quality and reliability of the data-set. The first type of sentences in the FASSILA corpus were derived from the Khouja corpus and translated into the Algerian Dialect by GPT-4 and verified by two Algerian native speakers. These sentences were already fact-checked by the owner authors, thus an additional annotation still required is for the SA which was done by two Algerian native speakers. As a result, these sentences served as a solid foundation for our corpus and provided a reliable set of true and false statements.

The second type of sentences was manually created by native Algerian Dialect speakers, who composed sentences based on current events and happenings. These contributors produced real news sentences, which were grounded in known facts, and Fake News sentences, which were generated by creating the opposite meaning of the real news sentences. In addition, the sentiment value also was provided with each sentence based on the sentiment expressed inside the sentence. The manual creation process helped ensure that the sentences were contextually relevant and representative of the AD, while also providing a balanced distribution for FN detection task.

The third type of sentences was extracted from YouTube comments, and required a more rigorous fact-checking process, as their veracity was unknown. To ensure the quality and reliability of these sentences, three annotators were employed to label each sentence as either real or fake for FN detection and positive, negative or neutral for SA task. All annotators were native AD speakers with extensive knowledge of the local culture and current events, and were trained to label FN and SA using a set of guidelines and criteria.

the fourth type of sentences were the ones paraphrased from the already existing sentences in the corpus including YouTube comments and manually created, using GPT-4 that saves the context while changing almost the entire sentences.
After the annotation and labeling process for both FN detection and SA was completed, the FASSILA corpus was assembled. This corpus offers a diverse and representative sample of statements in the AD, making it an invaluable resource for developing and evaluating FN detection and SA algorithms.

\subsubsection{Labeling criteria}

\begin{itemize}
    \item \textbf{Source Credibility: }
    Annotators assessed the credibility of the sources from which the sentences were derived. Trusted news outlets, government websites, and established organizations were considered more reliable, while sources with a history of publishing unverified or misleading content were deemed less credible. Annotators were encouraged to consult multiple sources to verify the accuracy of a statement before assigning a label.
    \item \textbf{Logical Consistency:}
    for this criterion, rigorous scrutiny of logical consistency ensured the integrity of analyzed sentences. Annotators meticulously examined statements for contradictions and inconsistencies, aligning them with established facts and knowledge. They employed a systematic approach, comparing content with reliable sources and theories to maintain logical integrity. When encountering inconsistencies, annotators deliberated carefully, consulting authoritative references to discern errors or falsehoods. Clear logical discrepancies led to labeling sentences as false, recognizing logical coherence as essential for reliability. This approach upheld the trustworthiness of the analyzed data.
    \item \textbf{Temporal Consistency: }
    for this criterion, the assessment of temporal consistency played a crucial role in ensuring the accuracy and relevance of the sentences analyzed. Annotators were responsible for considering the temporal context of the statements, meticulously evaluating whether the events or claims mentioned in the sentences aligned with the timeline of known facts. This meticulous evaluation aimed to identify and flag any statements that appeared outdated, irrelevant, or inconsistent with the current context.
    
    Annotators employed a systematic approach to assess the temporal consistency of the sentences. They carefully scrutinized the timeline of events, facts, and developments relevant to the research domain. This involved consulting authoritative sources, examining historical records, and considering established knowledge within the field. By leveraging these resources, annotators were able to establish a reliable temporal framework against which the sentences were evaluated as shown in figure \ref{fig:timeline}.

\end{itemize}

\subsubsection{Cohen's Kappa}
We have calculated the Cohen's Kappa ~\cite{kappa_ref} which is a statistical coefficient measure used to assess the agreement between two raters or evaluators in categorical or nominal data. It quantifies the extent of agreement beyond what would be expected by chance alone. To do that, the labeling process was done by 3 native speakers, where each one of them was given 300 sentences out of the total 1,436 sentences gathered from YouTube, where they annotated both for FA and SA.

\[
\kappa = \frac{{p_o - p_e}}{{1 - p_e}}
\]

where:
\begin{align*}
\kappa & : \text{Cohen's Kappa coefficient} \\
p_o & : \text{Observed agreement} \\
p_e & : \text{Expected agreement}
\end{align*}



After calculating the inter-annotator agreement using Cohen's Kappa coefficient, we found a value of 0.6488 (64.88\%) for the SA task. The lower agreement for the SA task can be attributed to the presence of three labels and three annotators. Much of the disagreement stemmed from confusion between the "Positive" and "Neutral" labels, which was influenced by the way speakers presented their ideas. Another round of verification was applied to reduce the disagreement.\\

\section{Dataset characteristics}

The FASSILA Corpus comprises 10,087 sentences, encompassing a vocabulary of over 19.497 words in the Algerian dialect (AD). Within the corpus, 5,393 sentences pertain to real news while 4,694 relate to FN, with each sentence distinctly labeled as authentic or fabricated. Across the corpus, seven primary categories are delineated: politics, COVID-related news, and topics reflecting prevalent interests in Algerian discourse during the research period: sports, rumors on car prices, tourism, eCommerce, and car accidents.\\
Structured comprehensively, the corpus integrates five essential elements: news category, data source, textual content, and labeling identifying each sentence as either authentic or fake news, alongside Sentiment Analysis categorizing each sentence as Positive, Negative, or Neutral. This structured approach facilitates future development of the FASSILA Corpus, empowering users to refine its categorization or employ specific subcategories for targeted tasks. Notably, the corpus maintains a balanced distribution of real and fake content within each category and across the corpus as a whole. The distribution nuances are outlined in detail within the table \ref{table: data stats},  offering insight into the corpus's composition and facilitating its utilization for diverse research and analytical purposes.
\begin{table}[h!]
\centering
\begin{tabular}{cc}
\hline
\textbf{Statistic} & \textbf{Value} \\ \hline
News timeline & 2017 - 2023 \\
Vocabulary size (Unique words) & 19,497 \\
Number of categories (Fig. \ref{fig:category}) & 7 \\
Number of sources (Fig. \ref{fig:source}) & 5 \\
Number of words in MSA & 3,936 \\
Number of words in Latin (En/Fr) & 42 \\
Number of characters in Latin & 1,404 \\
Avg. sentence length (AD) & 9.465 \\
Shortest sentence & 2 \\
Longest sentence & 255 \\ \hline
\end{tabular}
\caption{Data statistics for FASSILA.}
\label{table: data stats}
\end{table}
\begin{figure*}[h]
\centering
\includegraphics[scale=0.25]{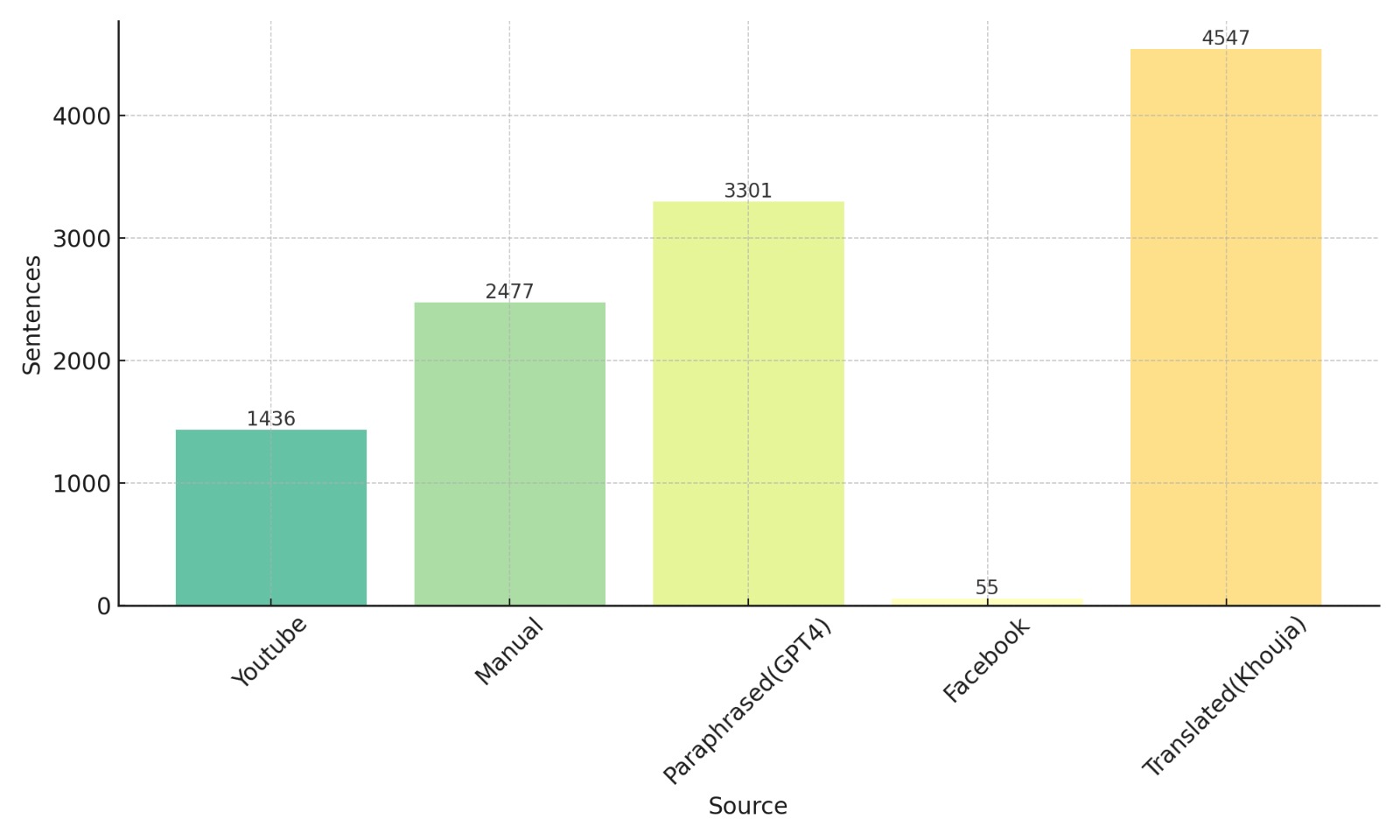}
\caption{Source diversity in FASSILA.}
\label{fig:source}
\                                     
\end{figure*}
\begin{figure*}[h]
\centering
\includegraphics[scale=0.25]{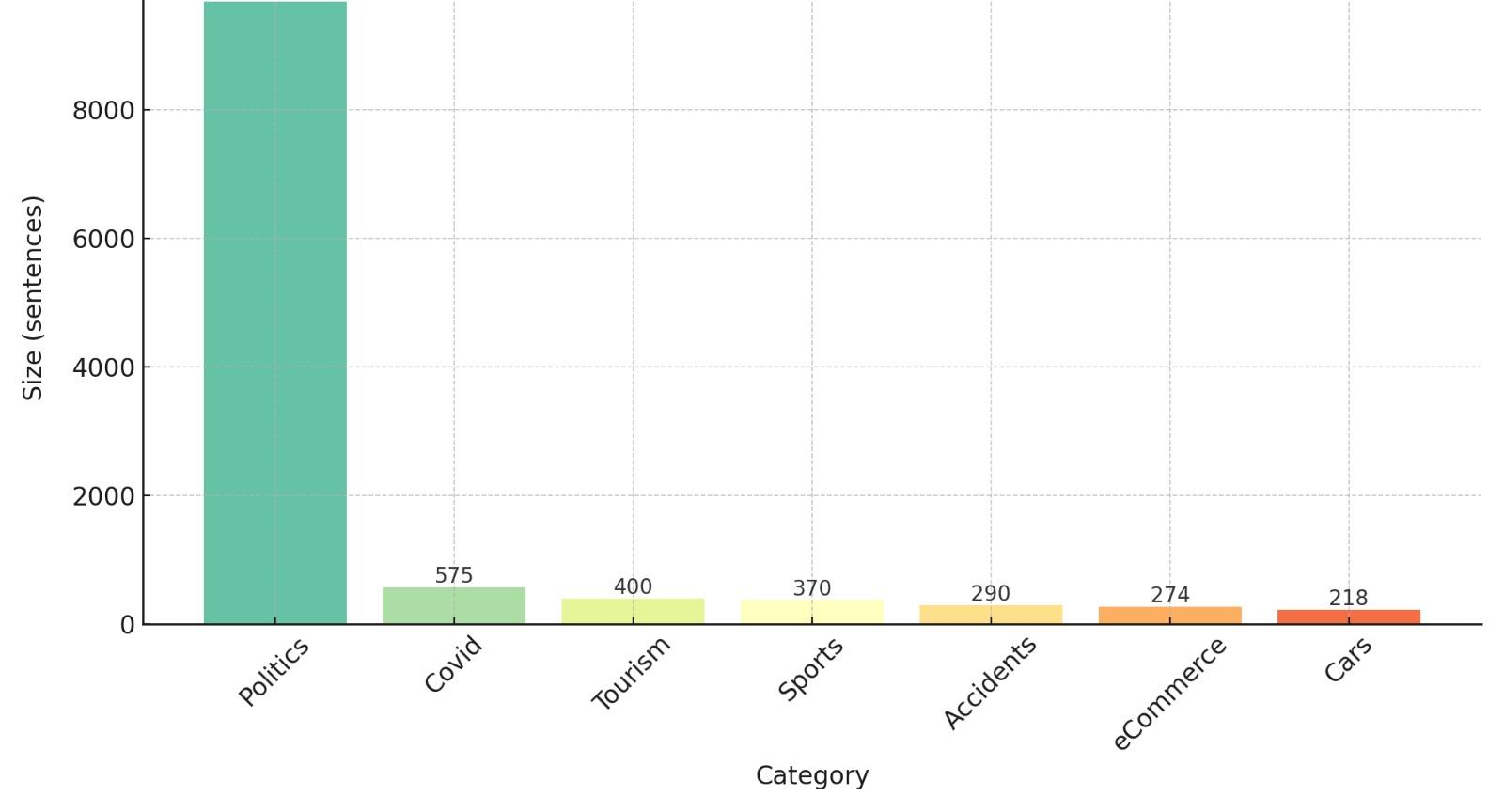}
\caption{Category diversity in FASSILA.}
\label{fig:category}
\                                     
\end{figure*}
Our corpus thoroughly encompasses the multifaceted nature of the AD, highlighting its sociolinguistic variations, distinctive phonetics and phonology, rich lexical mix, unique morphological features, and flexible syntax. These aspects are pivotal in understanding the complexity and diversity of AD, which is a part of the Maghrebi Arabic dialect continuum. For an in-depth comparison of Algerian Dialect with other linguistic styles, including a focus on its Arabic characters writing style, we refer to the detailed table provided in our study (see Table in \ref{tab:writing_styles}), which illustrates the variations between MSA and AD, among others.

\section{Experimental setup}

This section provides a comprehensive overview of our experimental framework. We detailing the transformer-based models and the Machine Learning baseline models that have been employed for the fake news detection task. Following this, we describe the evaluation metrics utilized. Additionally, we offer a thorough description of the dataset in terms of size and characteristics.

\subsection{Baselines}
As presented in Table \ref{models}, we employed three Arabic Pre-trained transformer-based models for our experiments: \textit{AraBERTv02}  ~\cite{AraBERTv2}, \textit{MarBERTv2}  ~\cite{MarBERTv2}, \textit{DziriBERT} ~\cite{DziriBERT} using colab pro, each chosen based on their training data and relevance to the dialects under study in our research. AraBERTv02 was selected for its predominant training on MSA and several dialects. \textit{MarBERTv2} was chosen due to its extensive training on the Maghrebi dialect, which encompasses Algerian, Moroccan, and Tunisian dialects.  \textit{DziriBERT} was opted for its focus on the Algerian Dialect. The comparison of these models provided a strong framework for examining the nuances and effectiveness of each in the particular context of identifying false news within the dialect in question. More details about these model is presented in Appendix \ref{model_details}.

\begin{table}[h!]
\centering
\begin{tabular}{llll}
\hline
\textbf{Model}     & \textbf{Params.} & \textbf{Tokens} & \textbf{Vocab.} \\\hline
AraBERTv02   & 136M       & 8.6B     & 64k         \\
\textit{MarBERTv2}   & 163M       & 6.2B     & 100k        \\
\textit{DziriBERT}     & 124M       & 20M     & 50k        \\
\hline
\end{tabular}
\caption{\label{models} The selected Arabic pre-trained models in terms of parameters number, size of training data (tokens), and the vocabulary size.}
\end{table}

In addition, a number of machine learning (ML) classifiers was utilized to analyze features extracted from the previously mentioned transformer-based models, which served as word embedding inputs. The primary objective of these classifiers was to predict the most appropriate category of this inputs. The range of classifiers deployed included, Support Vector Machine (SVM), K-Nearest Neighbors (KNN), Decision Tree (DT), Naïve Bayes (NB), and Logistic Regression (LR). It is imperative to emphasize that all ML algorithms were executed using their default parameter settings. In this way, we guaranty a uniform methodology across the various ML models, ensuring that any comparative analysis of their performance was based on their intrinsic algorithmic design rather than adjustments made through parameter tuning.

\subsection{Automatic Evaluation Metrics}

In our experimental framework, we have adopted a suite of metrics, including precision, recall, and F1-score, to rigorously evaluate the performance of the models in the context of binary text classification tasks. This meticulous selection of metrics ensures a well-rounded examination and subsequent analysis of the model’s capability to adeptly categorize the textual data into respective classes, aligning with the specificity and sensitivity requisites of our study.

\subsection{Dataset Description}

\begin{figure*}[ht]
\centering
\includegraphics[width=\textwidth]{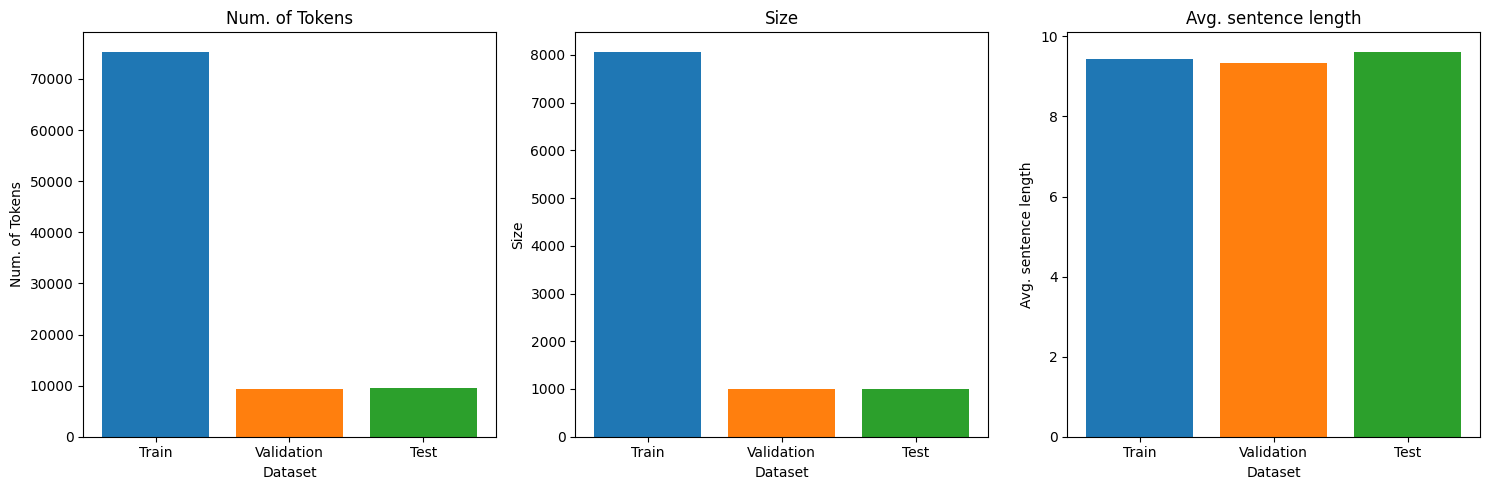}
\caption{Visualization of size, sentence length, and token count across training, validation, and test sets.}
\label{fig:visualization1}
\end{figure*}

\begin{figure*}[h!]
\centering
\includegraphics[width=\textwidth]{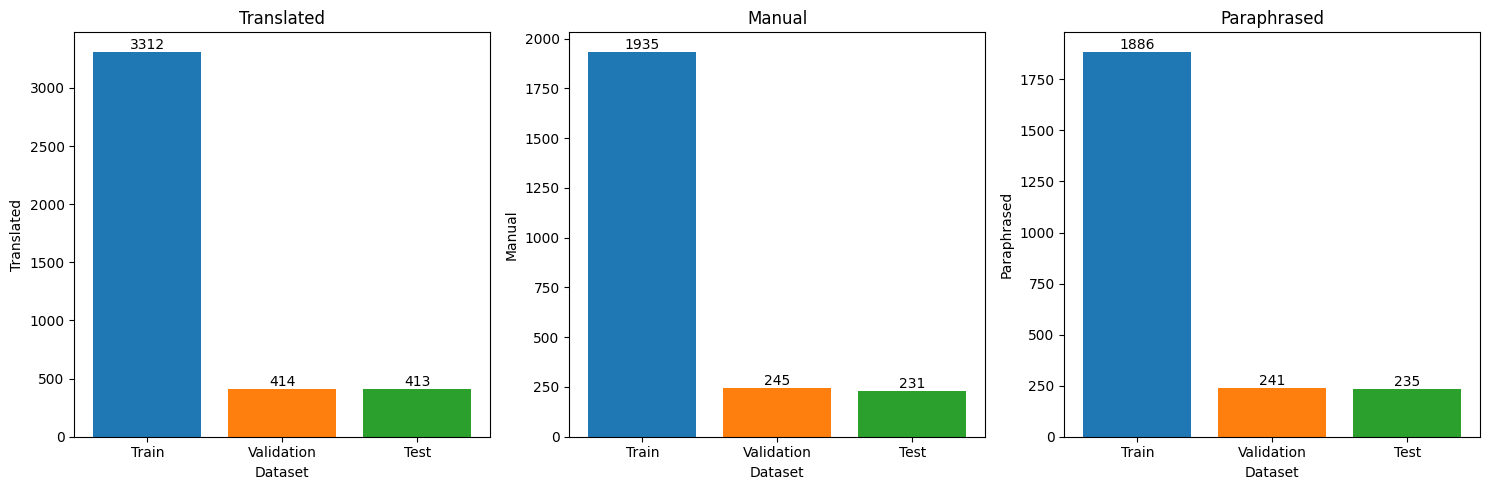}
\caption{Statistical overview by source for train, validation, and test sets.}
\label{fig:visualization2}
\end{figure*}

In these experiments, we addressed two tasks: fake news detection and sentiment analysis, using the FASSILA dataset. The experiments were structured as text classification tasks. For fake news detection, we utilized a binary format (fake and real), while for sentiment analysis, we employed three classes (positive, negative, and neutral). Statistical details about the dataset are provided in Table \ref{tab:dataset_statistics}. The visualizations in Figures \ref{fig:visualization1} and \ref{fig:visualization2} depict the numerical distribution and the most frequent data sources in the training subsets, respectively. 

\begin{table}[ht]
\centering
\begin{tabular}{lccc}
\hline
\textbf{Statistics}             & \textbf{Train} & \textbf{Validation} & \textbf{Test} \\ \hline
Num. of Tokens               &     75353               & 9318                        &    9604               \\ 
Size            &       8069             &      1009                   &       1009            \\ 
Avg. sentence length           &      9.436              &    9.340                     &      9.620             \\ 
Fake/Real        &     3755/4314               &    469/540                     &     470/539              \\ \hline
\end{tabular}
\caption{Statistics of the different training subsets.}
\label{tab:dataset_statistics}
\end{table}

\section{Results and Discussion}
To validate our corpus and showcase its strengths and potentials, we have done multiple tests and studies, and this section provide those results.

\subsection{Transformer-based Model Performance}
The results of analyzing the performance of transformer-based models Arabertv02, DziriBERT, and MARBERT on Fake News and Sentiment Analysis validation and test sets, presented in Table \ref{table:results_T}.

\begin{table}[h]
\centering
\caption{Comparative performance results of transformer-based models on Fake News test set.}
\label{table:results_T}
\begin{tabular}{@{}lccc@{}}
\toprule
& \textbf{Arabertv02} & \textbf{DziriBERT} & \textbf{MARBERT} \\ 
\midrule
\multicolumn{4}{c}{Test Set} \\
\cmidrule(r){2-4}
Accuracy       & 0.7799 & 0.774 & 0.7898 \\
Precision      & 0.7878 & 0.7484 & 0.7739 \\
Recall         & 0.7206 & 0.7739 & 0.7739 \\
F1-score       & 0.7527 & 0.761 & 0.7739 \\
\bottomrule
\end{tabular}
\end{table}

\begin{table}[h]
\centering
\caption{Comparative performance results of transformer-based models on Sentiment Analysis test set.}
\label{table:results}
\begin{tabular}{@{}lccc@{}}
\toprule
& \textbf{Arabertv02} & \textbf{DziriBERT} & \textbf{MARBERT} \\ 
\midrule
\multicolumn{4}{c}{Test Set} \\
\cmidrule(r){2-4}
Accuracy       & 0.8392 & 0.8065 & 0.8343 \\
Precision      & 0.7819 & 0.7929 & 0.7729 \\
Recall         & 0.8303 & 0.69 & 0.8023 \\
F1-score       & 0.8035 & 0.729 & 0.7864 \\
\bottomrule
\end{tabular}
\end{table}

MARBERT exhibits the highest accuracy on both sets, suggesting its superior capability in identifying fake news in this context due to its vocabulary size and the seen of Algerian dialect in the pre-training phase.In addition, MARBERT, not only is the accuracy highest, but its precision and F1-scores also outperform the other models. This indicates a balanced performance in terms of both false positives and false negatives.
Interestingly, DziriBERT shows the highest recall in the test set, which means it's better at identifying actual instances of fake news, albeit with slightly more false positives (as indicated by lower precision compared to MARBERT). Arabertv02 demonstrates consistent precision performance across validation, indicating reliable identification of true fake news cases which differ for others due to the large quantity of MSA seen during training which is different that the Algerian.

\subsection{ML Model Performance}

Tables \ref{table:arabertv02_results}, \ref{table:dziribert_results}, and \ref{table:marbertv2_results}, present the analysis of ML models' performance using fine-tuned embeddings from AraBERTv02, MARBERTv2, and DziriBERT. The non fine-tuned embedding results are presented in \ref{app:ML_non}.

\begin{table}[h!]
\centering
\caption{Test set performance results for ML models using fine-tuned AraBERTv02 embedding.}
\label{table:arabertv02_results}
\begin{tabular}{@{}lccc|ccc@{}}
\toprule
& \multicolumn{3}{c|}{Test FN detection} & \multicolumn{3}{c}{Test SA} \\
\cmidrule(lr){2-4} \cmidrule(l){5-7}
& KNN & SVM & LR & KNN & SVM & LR \\
\midrule
Accuracy   & 0.7809 & 0.775 & 0.771 & 0.8422 & 0.8442 & 0.8392 \\
Precision  & 0.7818 & 0.7676 & 0.7668 & 0.7929 & 0.7982 & 0.7921 \\
Recall     & 0.7334 & 0.7398 & 0.7292 & 0.8223 & 0.8222 & 0.8049 \\
F1-score   & 0.7568 & 0.7535 & 0.7475 & 0.8066 & 0.8096 & 0.7983 \\
\bottomrule
\end{tabular}
\end{table}

With AraBERTv02 embedding, The KNN model shows the highest accurac and F1-score on the validation set. However, in the test set, the differences between KNN, SVM, and LR are marginal, indicating a relatively consistent performance across these models with AraBERTv02.

Overall, the SVM model shows a consistent performance across different embeddings, particularly with MARBERTv2 and DziriBERT, while the KNN model stands out in the validation set using AraBERTv02. The choice of the embedding and model would depend on the specific balance of metrics (accuracy, precision, recall, F1-score) desired for the application.

\begin{table}[h!]
\centering
\caption{Test set performance results for ML models using fine-tuned MARBERTv2 embedding.}
\label{table:marbertv2_results}
\begin{tabular}{@{}lccc|ccc@{}}
\toprule
& \multicolumn{3}{c|}{Test FN detection} & \multicolumn{3}{c}{Test SA} \\
\cmidrule(lr){2-4} \cmidrule(l){5-7}
& KNN & SVM & LR & KNN & SVM & LR \\
\midrule
Accuracy   & 0.7908 & 0.7928 & 0.7888 & 0.0.8234 & 0.8293 & 0.8273 \\
Precision  & 0.778 & 0.7813 & 0.7844 & 0.7568 & 0.7771 & 0.7736 \\
Recall     & 0.7697 & 0.7697 & 0.7526 & 0.7898 & 0.7924 & 0.7857 \\
F1-score   & 0.7738 & 0.7755 & 0.7682 & 0.772 & 0.7843 & 0.7795 \\
\bottomrule
\end{tabular}
\end{table}

For MARBERTv2 embedding, we can observe that the performance is generally higher compared to AraBERTv02 which is the same observed with the fine-tuning process above. KNN and SVM models show almost similar accuracy in the test set, with a slightly higher precision and F1-score for SVM due to its complexity process.

\begin{table}[h!]
\centering
\caption{Test set performance results for ML models using fine-tuned DziriBERT embedding.}
\label{table:dziribert_results}
\begin{tabular}{@{}lccc|ccc@{}}
\toprule
& \multicolumn{3}{c|}{Test FN detection} & \multicolumn{3}{c}{Test SA} \\
\cmidrule(lr){2-4} \cmidrule(l){5-7}
& KNN & SVM & LR & KNN & SVM & LR \\
\midrule
Accuracy   & 0.7423 & 0.7621 & 0.7393 & 0.8115 & 0.8055 & 0.8095 \\
Precision  & 0.7154 & 0.7483 & 0.7219 & 0.7761 & 0.7727 & 0.7646 \\
Recall     & 0.7398 & 0.7356 & 0.7142 & 0.7141 & 0.6996 & 0.713 \\
F1-score   & 0.7274 & 0.7419 & 0.7181 & 0.7405 & 0.7294 & 0.7355 \\
\bottomrule
\end{tabular}
\end{table}

Laslty, The models using DziriBERT embedding demonstrate lower overall performance compared to MARBERTv2 but show improvement over AraBERTv02 in certain aspects, like SVM's accuracy in the test set.

\section{Conclusion}
\label{sec:conclusion}

The primary focus of our study was the construction of a dedicated corpus for the Algerian dialect, a critical achievement in the field of Natural language processing. This corpus is specifically tailored to the needs of Fake news detection and Sentiment analysis. We recognize that building this corpus is a challenging yet pivotal endeavor, as it serves as the foundation for the development of NLP tools, especially for low resources languages.

Our exploration of machine learning models underscored the importance of the corpus. It showed that choosing the right approach for binary classification tasks is critical, emphasizing the corpus's central role in NLP. Fine-tuning a pre-trained models based on transformer architecture, we experienced the direct impact of the corpus's construction. The varying performance of models, influenced by their training dialects and the presence of paraphrased sentences leaning towards the Moroccan dialect, emphasizes that the corpus's construction is at the core of Natural language processing success.

The FASSILA corpus is a innovative in Natural language processing for the Algerian dialect and it proven to be powerful when pared with transformer models and easy to finetune ML and DL models alike, serving as the foundational linguistic resource for Fake news detection and Sentiment analysis. Model selection should consider the dialectal diversity and dataset characteristics. Looking forward, the most critical avenue for further advancements lies in the continued construction and expansion of this valuable corpus. Addressing data imbalances, refining annotations, and exploring advanced model architectures are all steps that will further elevate the capabilities of linguistic resources for the Algerian dialect in the realm of Arabic NLP.

\bibliographystyle{unsrtnat}
\bibliography{main}
\appendix
\section{Appendix}
\label{sec:appendix}
\subsection{Hyperparameter choice}
\label{sec:hyperparams}
For AraBERTv02, we adopt best practices with a learning rate of $1,84 \times 10^{-5}$, $9$ epochs, batch size of $8$, weight decay of $0.01$, and without modifying the sequence length.
For MARBERTv2, we did the same as previous, with a learning rate of $2,95 \times 10^{-5}$, $10$ epochs, batch size of $8$, weight decay of $0.01$, and without modifying the sequence length. 
Similarly, DziriBERT also passed by hyperparameters tuning,  with a learning rate of $4,87 \times 10^{-5}$, $5$ epochs, batch size of $8$, weight decay of $0.01$, and without modifying the sequence length. 

\begin{table}[htbp]
    \centering
    \small
    \caption{Hyperparameter settings for our language models and KNN model.}
    \label{tab:hyperparameters}
    \begin{tabular}{lccc}
        \toprule
        Hyperparameter & AraBERTv02 & MARBERTv2 & DziriBERT \\
        \midrule
        Learning rate & 1.84 $\times 10^{-5}$ & 2.95 $\times 10^{-5}$ & 4.87 $\times 10^{-5}$ \\
        Epochs & 9 & 10 & 5 \\
        Batch size & 8 & 8 & 8 \\
        Weight decay & 0.01 & 0.01 & 0.01 \\
        K neighbor & 11 & 18 & 20 \\
        seed & 11 & 18 & 20 \\
        \bottomrule
    \end{tabular}
\end{table}

The fine-tuning and testing of models, was executed on the Google Colab platform, utilizing a Tesla T4 - 16GB GPU to harness optimal computational efficiency. \\Hyper-parameters were meticulously fine-tuned leveraging the validation set to ensure model efficacy. 
Throughout all our experiments using transformer-based models, we utilized Pytorch and the Hugging-face Transformers library ~\cite{wolf2020transformers}.

\subsection{Baseline details} \label{model_details}

\begin{itemize}
    \item \textbf{AraBERT:} is a BERT pre-trained model was trained on around 77GB of Arabic text (8B words) that included Wikipedia Arabic dump, OSCAR corpus~\cite{ortiz-suarez-etal-2020-monolingual}, OSIAN Corpus~\cite{zeroual-etal-2019-osian}, Abu El-Khair Corpus~\cite{elkhair201615} and a large collection from Assafir newspaper articles. 
    \item \textbf{MARBERT:} A large pre-trained model trained and released by the UBC NLP team. The model used a collection of over 1B tweets 128GB of text (15.6B tokens) in combination with 61GB of MSA text (6.5B tokens) from publicly available collections.  
    \item \textbf{DziriBERT:} is the first Transformer-based Language Model pre-trained dedicated just for the Algerian Dialect. It can handle both Arabic and Latin characters in Algerian text content. Even though it was pre-trained on significantly less data (one million tweets), it achieves new state-of-the-art performance on Algerian text classification datasets.
\end{itemize}

\begin{table}[h]
\centering
\caption{Statistics of fine-tuning Arabertv02, DziriBERT, and MARBERT models.}
\label{table:execution}
\begin{tabular}{@{}lccc@{}}
\toprule
& \textbf{Arabertv02} & \textbf{DziriBERT} & \textbf{MARBERT} \\ 
\midrule
\multicolumn{4}{c}{Test Set} \\
\cmidrule(r){2-4}
Train time (s) & 497.1001 & 269.373 & 624.0299 \\
Evaluation time (s)  & 1.4169 & 1.4781 & 1.573 \\
Train loss     & 0.2727 & 0.2837 & 0.1971 \\
Evaluation loss      & 1.1616 & 1.2114 & 1.6701 \\
Global steps   & 9081 & 5045 & 10090 \\
\bottomrule
\end{tabular}
\end{table}

\subsection{ML models results} 
\label{app:ML_non}
\begin{table}[h!]
\centering
\caption{ML models performance results using AraBERTv02 embedding.}
\begin{tabular}{@{}lccc|ccc@{}}
\toprule
& \multicolumn{3}{c|}{Test FN detection} & \multicolumn{3}{c}{Test SA} \\
\cmidrule(lr){2-4} \cmidrule(l){5-7}
& KNN & SVM & LR & KNN & SVM & LR \\
\midrule
Accuracy   & 0.6372 & 0.6759 & 0.661 & 0.749 & 0.7767 & 0.7857 \\
Precision  & 0.6093 & 0.6666 & 0.6331 & 0.6881 & 0.7797 & 0.7245 \\
Recall     & 0.6119 & 0.6055 & 0.6439 & 0.6136 & 0.5851 & 0.7057 \\
F1-score   & 0.6106 & 0.6346 & 0.6384 & 0.6417 & 0.635 & 0.7131 \\
\bottomrule
\end{tabular}
\end{table}

\begin{table}[h!]
\centering
\caption{ML models performance results using MARBERTv2 embedding.}
\begin{tabular}{@{}lccc|ccc@{}}
\toprule
& \multicolumn{3}{c|}{Test FN detection} & \multicolumn{3}{c}{Test SA} \\
\cmidrule(lr){2-4} \cmidrule(l){5-7}
& KNN & SVM & LR & KNN & SVM & LR \\
\midrule
Accuracy   & 0.6273 & 0.6531 & 0.6342 & 0.6974 & 0.7569 & 0.7787 \\
Precision  & 0.6504 & 0.638 & 0.6091 & 0.6794 & 0.7981 & 0.7425 \\
Recall     & 0.4285 & 0.5863 & 0.5948 & 0.4895 & 0.5289 & 0.693 \\
F1-score   & 0.5167 & 0.6111 & 0.6019 & 0.5233 & 0.5729 & 0.7142 \\
\bottomrule
\end{tabular}
\end{table}

\begin{table}[h!]
\centering
\caption{ML models performance results using DziriBERT embedding.}
\begin{tabular}{@{}lccc|ccc@{}}
\toprule
& \multicolumn{3}{c|}{Test FN detection} & \multicolumn{3}{c}{Test SA} \\
\cmidrule(lr){2-4} \cmidrule(l){5-7}
& KNN & SVM & LR & KNN & SVM & LR \\
\midrule
Accuracy   & 0.6174 & 0.665 & 0.661 & 0.7311 & 0.7559 & 0.7271 \\
Precision  & 0.6004 & 0.6555 & 0.6348 & 0.6663 & 0.7899 & 0.6612 \\
Recall     & 0.5287 & 0.5884 & 0.6375 & 0.6038 & 0.5686 & 0.6306 \\
F1-score   & 0.5623 & 0.6202 & 0.6361 & 0.62857 & 0.6209 & 0.6435 \\
\bottomrule
\end{tabular}
\end{table}

\subsection{GPT-4 Prompts}\label{app:prompts}
In the below sections, we report the prompts used for translation and paraphrasing tasks. GPT-4 was employed with default parameters, without limiting the output length.

\subsubsection{Translation}

To generate an automatic translation, we employed the following prompt \textit{"Please translate the following to the exact Algerian dialect, please don't confuse it with any Darija dialects such as Moroccan and Tunisian, and try your best please:"} through an API call, with the input being MSA text.
\subsubsection{Paraphrasing}
To generate paraphrased text, we employed the following prompt \textit{"Please paraphrase the following sentences while preserving the exact Algerian dialect while maintaining the context, please don't confuse it with any Darija dialects such as Moroccan and Tunisian, and try your best please:"} through an API call, with the input being Algerian text.

\section{Breakdown of translation process} \label{app:breakdownT}

For the translation task, we made a comparative research to pick the best machine translator that can held the translation from MSA to Algerian dialect, focusing on two prominent LLMs and our own custom model. We employed a zero-shot approach to evaluate the two LLMs and a fine-tuning approach for our custom model. This assessment encompasses both the automatic translation, utilizing metrics such as BLEU,  Chrf++, and COMET and a manual translation. The manual translation involves selecting the most appropriate model for each sentence, and take the model that achieved the highest score. This process undertaken by a panel of three Algerian experts in the field as a human assessment.

\subsubsection{\textbf{Baselines}}
\begin{itemize}
    \item \textbf{Nllb-200-distilled-600M} ~~\cite{costa2022no}: A multilingual neural machine translation system that encompasses 200 languages, based on the transformer encoder-decoder framework. For our purposes, we leveraged two versions of this model. One is tailored for translating MSA to Moroccan text, while the other for MSA to Tunisian. This choice was driven by the overlapping characteristics observed among these two dialects and Algerian dialects.
    
    \item \textbf{GPT-4}: The latest in the GPT series from OpenAI, this model stands out for its capability to comprehend and generate both natural language and code. We chose GPT-4 for its proficiency in handling less common languages and dialects, such as the Algerian dialect, for which specific translators are not readily available. Pre-trained on a vast corpus encompassing multiple languages and domains, and it's generally understood to be an auto-regressive language model with a foundation in the transformer architecture ~~\cite{vaswani2017attention}.  
    
    \item \textbf{Our custom model}: To address the gap of the no existence of MSA-Algerian translators,this model was created by optimizing the AraBART model ~~\cite{eddine2022arabart} to focus on converting MSA text to Algerian dialect. It was trained using a set of 11,722 parallel sentences from our curated and annotated dataset for another project.
\end{itemize}

\subsubsection{\textbf{Dataset}}
For this task, we used a widely recognized dataset is crucial. In this study, we've chosen the "MADAR CORPUS-25" ~~\cite{bouamor2018madar}, which consists of 2K sentences for each 25 distinct city dialects. Each sentence is paired with 25 parallel translations, including both the Algerian dialect and the MSA version. After thorough verification and data cleaning, we retained 1,588 parallel sentences as a test set.

\subsubsection{\textbf{Automatic evaluation}}

The automatic evaluation's findings shown in Table \ref{comparative}, reveal insightful performance distinctions. GPT-4 and NLLB-MOR exhibit superior performance across all three metrics: BLEU (9.42/8.94), Chrf++ (23.08/23.39), and COMET (66.74/67.71), indicating their effectiveness in capturing both the linguistic nuances and contextual alignment in translation.

\begin{table}[h]
\centering
\begin{tabular}{llll}
\hline
\textbf{Model} & \textbf{BLEU} & \textbf{Chrf++} & \textbf{COMET}\\
\hline
GPT-4  & { 9.42} &  23.08 & 66.74 \\
Our Custom Model      & { 5.95} & 19.07 & 60.82   \\
NLLB TUN       & { 8.22} & 21.30 & 65.34 \\
NLLB MOR       & { 8.94}  & 23.39  & 67.71\\
\hline
\end{tabular}
\caption{\label{comparative}Comparative analysis of different models on MSA-AD translation based on BLEU, Chrf++ and COMET.}
\end{table}

\subsubsection{\textbf{Manual evaluation}} On the other side, for the manual translation, we found that out of a total of 1,588 sentences, GPT-4 demonstrated the highest capability, effectively translating 775 sentences. This aligns with its superior performance observed in the automatic metrics. NLLB MOR, with its proficiency in 479 sentences, and our custom model, accurately translating 447 sentences, also reflected their respective standings in the earlier evaluation. NLLB TUN, while translating 412 sentences effectively, mirrored its slightly lower performance in the automatic metrics. These findings reinforce the correlation between the automatic evaluation results and expert human judgment in assessing translation quality.

In conclusion, for a sensitive task such as fake news detection, we prioritize GPT-4 model that accurately captures the context of the input sentence and preserves the same meaning of the sentence.







\end{document}